\documentclass[3p,times,procedia]{elsarticle}
\usepackage{ecrc}
\usepackage[bookmarks=false]{hyperref}
    \hypersetup{colorlinks,
      linkcolor=blue,
      citecolor=blue,
      urlcolor=blue}

\volume{00}

\firstpage{1}

\journalname{Procedia Computer Science}

\runauth{Mohamad Ballout et al.}

\jid{procs}

\usepackage{amssymb}

\usepackage[figuresright]{rotating}

\begin{document}
\begin{frontmatter}



\dochead{International Neural Network Society Workshop on Deep Learning Innovations and Applications (INNS DLIA 2023)}

\title{Investigating Pre-trained Language Models on Cross-Domain Datasets, a Step Closer to General AI}


\author{Mohamad Ballout\corref{cor1}} 
\author{Ulf Krumnack}
\author{Gunther Heidemann}
\author{Kai-Uwe Kühnberger}

\address{Institute of Cognitive Science, University of Osnabrück, Osnabrück, Germany}

\begin{abstract}
Pre-trained language models have recently emerged as a powerful tool for fine-tuning a variety of language tasks. Ideally, when models are pre-trained on large amount of data, they are expected to gain implicit knowledge. In this paper, we investigate the ability of pre-trained language models to generalize to different non-language tasks. In particular, we test them on tasks from different domains such as computer vision, reasoning on hierarchical data, and protein fold prediction. The four pre-trained models that we used, T5, BART, BERT, and GPT-2 achieve outstanding results. They all have similar performance and they outperform transformers that are trained from scratch by a large margin. For instance, pre-trained language models perform better on the Listops dataset, with an average accuracy of 58.7\%, compared to transformers trained from scratch, which have an average accuracy of 29.0\%. The significant improvement demonstrated across three types of datasets suggests that pre-training on language helps the models to acquire general knowledge, bringing us a step closer to general AI. We also showed that reducing the number of parameters in pre-trained language models does not have a great impact as the performance drops slightly when using T5-Small instead of T5-Base. In fact, when using only 2\% of the parameters, we achieved a great improvement compared to training from scratch. Finally, in contrast to prior work, we find out that using pre-trained embeddings for the input layer is necessary to achieve the desired results.
\end{abstract}

\cortext[cor1]{Corresponding author. Tel.: +49 541 969-3384}
\end{frontmatter}

\email{mohamad.ballout@uni-osnabrueck.de}


\vspace*{-6pt}
\section{Introduction}
\label{intro}

The ultimate goal of artificial intelligence is to achieve general intelligence where the machine is able to simulate human cognitive abilities. In theory, a general AI model has a wide range of capabilities and can solve a diverse set of problems, rather than being designed to perform a specific task. Thus, in this paper, we investigate whether pre-trained language models can perform well and improve the state-of-the-art on diverse tasks that are not directly related to language. This could imply that the models have gained general knowledge during pre-training that is useful on downstream tasks from different domains. 

We chose to investigate transformers because since they were proposed \cite{b1}, they replaced RNNs in most NLP tasks such as translation, language modeling, sentiment analysis etc. They were also applied to other domains including vision \cite{b2} and audio processing \cite{b3} where they also achieve state-of-the-art performance.

 Pre-training transformers on large corpora have emerged as a standard practice before fine-tuning on downstream tasks. The architecture of these pre-trained language models varies between the standard encoder-decoder architecture such as T5 \cite{b4} and BART \cite{b5}, an encoder only architecture like BERT \cite{b6}, and a decoder only architecture like GPT \cite{b7}. 
Most of the prior work employ pre-trained models by fine-tuning them on specific NLP tasks. However, in this paper, we investigate whether pre-training models on language can help on tasks from different domains and hence test their ability to generalize. Pre-training on language is attractive as it can be performed in an unsupervised fashion with unlabeled text data, abundant on the web.  Pre-training on language would be more important if models were enabled to generalize to tasks from different domains. Thus, we choose four well-known pre-trained transformers with different architectures including T5, BART, BERT, and GPT-2 and we fine-tune them on three different non-language tasks. Two tasks are taken from the Long Range Arena Dataset \cite{b8}, which are Long Listops and CIFAR10-LRA, and the third is Remote Homology by TAPE \cite{b9}. We compare the performance of our pre-trained models to transformers trained from scratch from \cite{b8}.

The number of parameters in pre-trained models varies from millions to billions of parameters. For instance, the T5 model was released in five settings, small, base, large, 3B, and 11B. The number of parameters spans from around 60 millions for the small model to 11 billion parameters for the largest model. The models are usually extended by increasing the embedding size, the number of head-attention, the number of encoder-decoder layers, and the size of the feed forward layers. In this paper, we investigate the effect of the size of the model by using both the small and the base model of T5.

Our findings suggest that using pre-trained language models increases the accuracy by a large margin. We also use the method suggested by \cite{b10} in which we freeze 98\% of the parameters and the performance remains rather good but slightly lower. These results suggest that during pre-training models develop general knowledge that could be useful for tasks that are not related to language such as vision tasks or tasks that require reasoning on hierarchical data structures. Unlike prior work that suggest to reinitialize the input layer, we show that using a pre-trained input layer is crucial to get the desired results.

The main contributions of this paper are:
\begin{itemize}
\item
We investigate how well pre-trained language models generalize to datasets from different domains. We achieve a large improvement by using models like T5 that scores 64.2\% on the Long Listops dataset and 58.5\% on CIFAR10-LRA and 14.2\% on Remote Homology, whereas the state-of-the-art transformers accuracy are around 38\%, 40\% and 13\%, respectively. Based on the goal that each dataset was created for, we claim that pre-training on language helps models to acquire general skills that helped in improving the accuracy on these tasks. For instance, the improvement on Listops, which is a hierarchical task, shows that the pre-trained models learned reasoning on hierarchical data structures during pre-training on language. 
\item
We compare four well-known pre-trained language models including T5, BART, BERT, and GPT-2. We show that all of these models have a similar superior performance, showing a good capability of generalizing to different tasks.  
\item
Unlike prior work, where re-initializing the input of pre-trained models is a standard practice when reading data from a new domain, we show that using pre-trained embeddings is crucial to achieve the improvement in accuracy. 
\item
Finally, we show that a good performance can be achieved even when decreasing the number of parameters from 220 to 2 million when freezing 98\% of T5-Small parameters.
\end{itemize}

\section{Related Work}

Transfer learning with transformers started by training models like T5\cite{b4}, BART\cite{b5}, BERT\cite{b6}, GPT\cite{b7}, and Roberta\cite{b11} in an unsupervised manner on large corpora and fine-tuning them on downstream tasks. It was found that training on large corpora enables the network to learn language representations that are extremely useful in downstream language tasks such as spam detection \cite{b133}, sentiment analysis \cite{b144}, translation \cite{b4}, etc.

Similarly, in the vision domain, transformers  are trained on large datasets such as ImageNet-21k \cite{b12} or JFT-300M. Models like ViT \cite{b2} and GPT pixels \cite{b13} are then fine-tuned on downstream vision tasks and they achieve state-of-the-art performance. Thus, the standard practice is to pre-train and then fine-tune models on tasks from the same domain. In this paper, we investigate if we can find universal pre-training data that is useful on various tasks. Thus, we investigate whether pre-training on language help the models to develop general knowledge by fine-tuning them on datasets from different domains.  

This work was inspired by \cite{b14} where they use T5, a pre-trained language model, to sum two 60-digit numbers. They fine-tune a pre-trained language model on an arithmetic task. However, their goal was to study the importance of the surface representation of the input. They re-represent the input and show that explicitly representing the decimal place of digits in long numbers helps the model to do the sum correctly. They did not compare their results with trained from scratch transformers, but their work inspired us to use T5 on the arithmetic dataset Long Listops and furthermore CIFAR10-LRA and Remote Homology.

Our work is similar to prior work investigating unified models for multi-task learning \cite{b15, b17}. Multi-task learning is the approach of training a single model that can perform well on multiple tasks at once. Our work focuses on investigating the effect of pre-training on non-language tasks, whereas the previous efforts try to build architectures to solve multiple tasks. Most of the multi-task architectures in previous work use common parameters for all of the tasks, but some of them also have parameters that are specific for each task. For example, decoders in \cite{b17} and heads in \cite{b15} are specifically designed for each task.

Our work is most similar to frozen pre-trained transformer (FPT) \cite{b10} with some major differences. In their paper, the authors use a pre-trained GPT-2 model but with freezing most of the parameters (99.9\%). They fine-tune the remaining parameters on tasks from different domains including numerical computation, vision, and protein fold prediction. They show that even when training 0.1\% of the parameters, GPT-2 was able to achieve a similar or a slightly better performance compared to training a transformer from scratch on non-language tasks. They also showed that fine-tuning all of the parameters in the pre-trained model does not help, but in contrast the accuracy diverges. Their fully trained model scores 35.8\% and 21.0\% on the Listops and CIFAR10-LRA \cite{b8}, whereas our GPT-2 scores 43.2\% and 62.0\%, respectively. The main difference is to use a pre-trained input layer instead of re-initializing it randomly. In addition, the authors \cite{b10} try GPT-2 as their main pre-trained model, whereas we do a more extensive study with four different pre-trained models including GPT-2, BART, BERT, and T5. All of these models achieve similar results with one exception discussed in the results section. We achieve a better performance than FPT and transformers trained from scratch when training 100\% and 2\% of the models parameters.

\section{Methodology}

\subsection{Multi-Modal Tasks}
\label{sec: 3 1}
In order to test our hypothesis, we fine-tune the four pre-trained language models on three datasets from different domains:

\begin{itemize}
\item
\textbf{Long Listops}: The Long Listops dataset is a classification task taken from the Long Range Arena Dataset \cite{b8}. It is a longer version of the original Listops dataset \cite{b18}. This dataset is made to test the capability of modeling hierarchical data in a long setup. It contains mathematical operations including Modular sum, Minimum, Maximum, and Median. A short sample of a sequence could be: ``\textit{ [MAX 0 [MIN 1 4] 3 [SUM 2 7] ]} ". The correct output in that case would be 9. Prior work shows that transfomers struggle with this type of data as the model has to remember the order of every token to correctly predict the answer. It is a 10 classes classification with 96k training samples with a sequence length range from 500 to 2000 tokens.  
\item
\textbf{CIFAR-10 from LRA}: This dataset is also taken from Long Range Arena Dataset. It is a modified version of the original CIFAR-10 dataset \cite{b19} i.e. an image classification task including 10 classes. It is considered to be a more challenging task in comparison to the original version since an $N \times N$ image is flattened to one dimension of length $N^2$. There are 50k training samples and the images are converted to gray scale for simplicity. In this task, the model is required to learn 2D patterns for an input of one dimension.  

\item
\textbf{Remote Homology Detection}: This is a classification task taken from Tape \cite{b9}. The input is a sequence of amino acids and the label is its protein fold. A sample of the amino acids sequence is: ``KSPLTYAEALANTIMNTELPPANR..GK". Each letter in this sequence refers to an amino acid and they are encoded using the standard 25-character alphabet. For instance, ``A" refers to Alanine and ``G" to Glycine. The length of this sequence is around 400 amino acids. The length of the sequences in the dataset spans from around 20 to more than 1000. Given its amino acids sequence, the model should predict the protein fold, which is the specific three-dimensional structure represented in the dataset by 1195 different classes, that a protein adopts. This task tests the capability of a model to find structural similarities from inputs that are not directly related. It is considerably a challenging task given that only around 12k training samples are given for 1195 classes.

\end{itemize}

As shown above, we chose a variety of datasets from different domains. We are investigating whether models acquired some general abilities during pre-training that could help in these tasks. For instance, Listops was originally created by \cite{b18} to simulate parsing in language. We investigate whether pre-training on language might have taught the models the ability of parsing. In fact, investigating Listops would also lead us to detect whether the models have the ability of  reasoning on structured data since it has sequences with mathematical operators which are enclosed with brackets.

On the other hand, even though CIFAR10-LRA is a vision task where the model has to do image classification which is considered the farthest from language, the task still captures whether models can learn relations between 2D pixels while represented in one dimension. Thus, we investigate if pre-trained language models could make use of learning relations between words and project it to learning relations between pixels. Similarly, the Remote Homology dataset investigates whether pre-training gives models the ability to generalize since the task requires the model to detect structural similarities across distantly related inputs.

\subsection{Flowchart of the Investigation}

\begin{figure}[htbp]
\begin{center}
\includegraphics[scale=0.6]{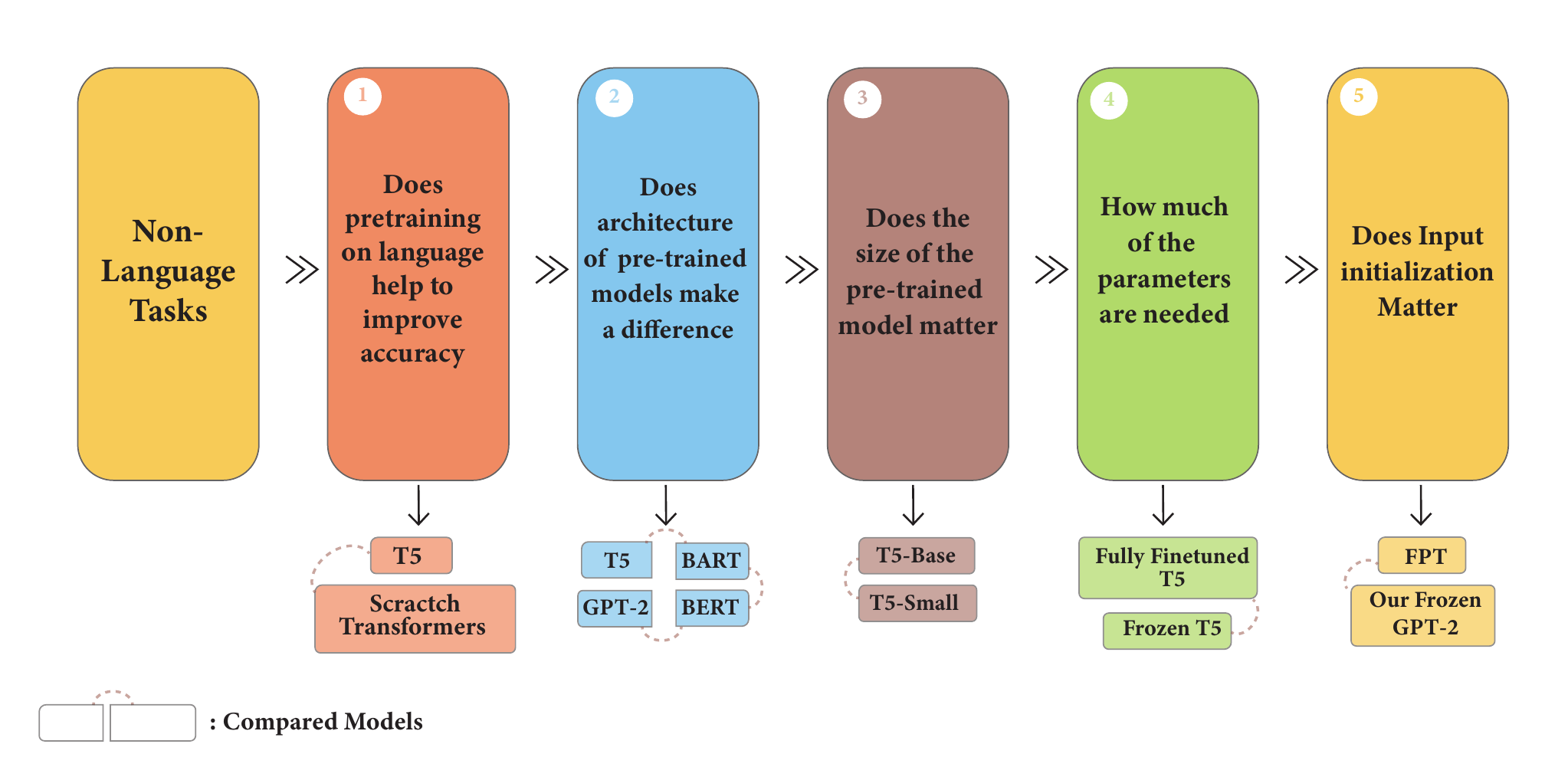} 
\caption{Figure presents our ablation study, assessing language pre-training effects on diverse tasks}
\label{fig:2}
\end{center}
\end{figure}

Since the main goal of this work is to investigate the effect of language pre-training on downstream tasks from different domains, we do an ablation study to find the role of each step shown in figure \ref{fig:2}. 

First, we study the effect of language pre-training on different modalities by comparing one pre-trained model, T5, with 10 state-of-the-art trained from scratch transfomers from \cite{b8}, which includes models like Big Bird \cite{b20}, Longformer \cite{b21}, Linformer \cite{b22}, etc.

Second, we compare four well-reputable pre-trained models including T5-Base, BERT, GPT-2, and BART. We chose a diversity of models including two encoder-decoder models BART and T5-Base, an encoder model, BERT, and a decoder model, GPT-2. We investigate whether the architecture of the model has any effect on the results and to ensure that the improvement is from pre-training regardless of the architectures or the pre-training techniques. All of these models are pre-trained on large language corpora with different techniques. For instance, BERT is designed to have deep bidirectional representations, whereas GPT-2 is trained as a left-to-right generative model. BART is a combination of BERT as an encoder and GPT as a decoder. T5 is an encoder-decoder model that has a similar architecture to the original transformer \cite{b1}. In the T5 paper, the authors study the best practices for transfer learning and they apply it to their model to get state-of-the-art results.  
After comparing four different architectures, in step 3, we compare T5-Base with T5-Small in order to study the effect of the size of the models. T5-Small is a scaled down version of T5-Base but with a similar architecture and training strategy. 

Additionally, we also try fine-tuning the four suggested pre-trained models in two settings. In one setting, we fine-tune all of the parameters of the models, whereas in the second setting, we freeze 98\% of the parameters and train only the layer norm parameters and the output layer which makes around 2\% of the trainable parameters. This allows for models with less trainable parameters and fast fine-tuning. 

Finally, we compare our results with FPT from \cite{b10} where they froze most of the parameters and they reinitialize their input layer. It is a standard practice to reinitialize the input layer when reading a new modality. However, since our goal is to investigate the effect of language pre-training, we initialized our input from the pre-trained embeddings. We show that this initialization has a great effect on the performance of the model. Figure \ref{fig:2} summarizes the questions to be investigated in this paper and the models compared to answer these questions.

\section{Results and Discussions}

\subsection{Does pre-training on language help on non-language tasks?}
\label{sec: 4 1}
The first step in our investigation is to compare a pre-trained model, which we choose to be the T5 encoder-decoder model. T5 is pre-trained on large corpora that they called ``Colossal Clean Crawled Corpus''. The base model of T5 has around 220 million parameters that are fine-tuned with a learning rate of 3e-4. The remaining hyper-parameters are adopted from \cite{b14}. In table \ref{tab:t5 vs scratch}, we are comparing the pre-trained model T5 with the top performance transformer trained from scratch on each dataset from \cite{b8}. We can see that T5 got an accuracy of 64.2\%, while the best transformer which is Reformer \cite{b23} scored 37.3\%. In addition, T5 scored on CIFAR10-LRA 58.5\%, whereas the best performance was 44.2\% by Sparse Transformer \cite{b24}. These results show that even though pre-trained on language data, T5 improved the accuracy from 37.3\% to 64.2\% on Listops which is an arithmetic dataset and from 44.2\% to 58.5\% on CIFAR10-LRA which is an image classification task. It is important to emphasize that this state-of-the-art performance is on CIFAR10-LRA which is not to be confused with CIFAR-10 where the current performance is as high as 99.5\%.

These results hint that pre-training on language helps the model to achieve a better performance on tasks from different domains. The large improvement could be attributed to the fact that the model acquired general knowledge during pre-training on language. For instance, the large improvement on a highly hierarchical dataset such as Listops hints that during pre-training on language, the model was able to acquire reasoning on hierarchical data that was retrieved during fine-tuning. Similarly, the pre-trained model proved that it was able to map the learned relation between words to relation between pixels and it was able to improve the accuracy on CIFAR10-LRA significantly. Overall, the improvement on both non-language datasets suggests that the pre-trained language model had acquired implicit general reasoning during pre-training which is useful on non-language tasks.           

\begin{table}[htbp]
\caption{Results of Pre-trained model T5 versus transformers trained from scratch on Listops and CIFAR10-LRA}
\begin{tabular*}{\hsize}{@{\extracolsep{\fill}}lll@{}}
\toprule
Model & Listops & CIFAR10-LRA\\
\colrule
T5 & \textbf{64.2\%} &  \textbf{58.5\%}\\
Scratch Trans. & 37.3\%  &  44.2\% \\
\botrule
\label{tab:t5 vs scratch}
\end{tabular*}
\end{table}

\subsection{Does the architecture of pre-trained models make a difference?}
In this section we further investigate the large improvement shown in \ref{sec: 4 1}. In order to ensure that the improvement is due to pre-training on language and not due to the features of the architectures of the respective model, we test four diverse pre-trained models instantiating different architectures including T5, BART, BERT, and GPT-2. Table \ref{tab 2:parameters of models} shows the number of parameters, the type, and the training strategy of each model.

\begin{table}[htbp]
\caption{Table shows the four different pre-trained models used in our investigation and their characteristics}
\begin{tabular*}{\hsize}{@{\extracolsep{\fill}}llll@{}}
\toprule
\textbf{Model}& \textbf{Num.\ of Parameters} & \textbf{Architecture} & \textbf{Training Strategy} \\
\colrule
T5  & 220M &  Encoder-Decoder & Bidirectional  \\

BART & 140M  &  Encoder-Decoder  & Bidire + Auto-Regr. \\

BERT & 110M  &  Encoder Only & Bidirectional \\

GPT-2 & 125M  &  Decoder Only & Auto-Regressive\\
\botrule
\label{tab 2:parameters of models}
\end{tabular*}
\end{table}

We investigate four diverse pre-trained language models with different training strategies and different sizes. T5 and BERT are both bidirectional models where the representation of the words are derived from both its left and right context. The difference between T5 and BERT in the training strategy is that BERT masks a token for each word, whereas T5 replaces multiple consecutive tokens with a single mask. It is up to the model to figure out how many tokens are missing in the sentence. Also, T5 is similar to the original transformer in architecture. It has encoder and decoder, whereas BERT is an encoder only model. Thus, T5 has 220 million parameters, whereas BERT has only 110 millions. GPT-2 is an auto-regressive model and the representation of the word is only derived from the left context. It is known that this type of models usually do not perform well on tasks that require knowledge of the whole sequence. GPT-2 has around 125 million parameters. Finally, similar to T5, BART is an encoder-decoder model that combines both BERT and GPT-2 strategies by having a bidirectional encoder and auto-regressive decoder. It has around 140 million parameters. 

We test the four models on Listops and CIFAR10-LRA, but it is important to note that in this experiment and for simplicity we use Listops sequence length of 200-500 instead of 500-2000 used in section \ref{sec: 4 1}. Table \ref{tab 3:pre-trained models} shows that all of the pre-trained models score higher than the best performing transformer trained from scratch. The performance of the pre-trained language models is similar where T5 scores the highest on Listops(200-500) with an accuracy of 65.4\% compared to 39.5\% for the best transformer trained from scratch. BART scores the highest on CIFAR10-LRA with an accuracy of 62.1\%. However, it is shown that all of the pre-trained models have a similar performance except for GPT-2 on Listops that scores 43.2\%, which is significantly lower than other pre-trained models. As stated before, it is usually not recommended to use auto-regressive models for a classification tasks, even though the model performs well on CIFAR10-LRA.

\begin{table}[htbp]
\caption{Table showing the accuracy of the four pre-trained models on  Listops and CIFAR10-LRA}
\begin{tabular*}{\hsize}{@{\extracolsep{\fill}}lll@{}}
\toprule
\textbf{Model}& \textbf{Listops(200-500)} & \textbf{CIFAR10-LRA} \\
\colrule
T5  & \textbf{65.4\%} & 58.5\%  \\

BART & 62.5\%  &   \textbf{62.1\%} \\

BERT & 63.5\%  &  57.5\%  \\

GPT-2 & 43.2\%  &  62.0\% \\

Scratch Trans. & 39.5\%  &  44.2\% \\
\botrule
\label{tab 3:pre-trained models}
\end{tabular*}
\end{table}

An explanation of this performance is that Listops is a dataset where a model needs to remember the order of every token in the sequence to predict the answer correctly, whereas in the vision task CIFAR10-LRA a model can predict the answer correctly even when it misses some pixels. Finally, BERT performs well on both datasets supporting the convention that bidirectional models perform better on classification tasks.    

It is also important to note that the accuracy did not improve significantly when shortening the Listops sequences from 500-2000 to 200-500. The difference is only around 1\% as shown in table \ref{tab:t5 vs scratch} and \ref{tab 3:pre-trained models}. Based on this observation we compare the average of pre-trained language models with the average of models trained from scratch in table \ref{tab 4: Average accuracy}. We can see that  pre-trained language models improve the accuracy from 29.0\% to 58.7\% on the Listops dataset and from 41.0\% to 60.2\% on the CIFAR10-LRA dataset.  

\begin{table}[htbp]
\caption{Table showing the average accuracy of pre-trained models versus models trained from scratch}
\begin{tabular*}{\hsize}{@{\extracolsep{\fill}}lll@{}}
\toprule
\textbf{Models Type}& \textbf{Listops} & \textbf{CIFAR10-LRA} \\
\colrule
Pre-trained Avg & \textbf{58.7\%}  &  \textbf{60.2\%} \\
Scratch Trans. Avg & 29.0\%  &  41.0\% \\
\botrule
\label{tab 4: Average accuracy}
\end{tabular*}
\end{table}

The results shown in this section demonstrate that the improvement on the cross-domain datasets is independent of the architecture of the model and its training strategy. All of the pre-trained language models were able to significantly improve the accuracy compared with transformers trained from scratch regardless of whether the model is an encoder, a decoder, or an encoder-decoder and regardless of the training strategy. Even though the results show that both, architecture and training strategy might have a small impact, a major improvement was achieved by all of the models and the common factor was pre-training on language. The results in this section hint again that pre-training on language moves us a step closer to general AI that solves multi-task problems rather than being designed to perform a specific task in a specific domain.

\subsection{Does the size of pre-trained models matter?}

Since in the previous section most of the pre-trained transformers showed a similar performance and it was not clear whether the number of parameters in the model matters, in this section we compare T5-Base that has 220 million parameters with T5-Small that has around 60 million parameters. Table \ref{tab 55:t5 vs small t5 architecture} shows a detailed comparison between T5-Base and T5-Small. The embedding size of T5-Base is 768, whereas T5-Small uses 512. In addition, T5-Base uses 12 attention heads and 12 blocks while T5-Small uses 8 attention heads and 6 blocks. Each block consists of attention, add and norm, and feed-forward layers and the higher the number of blocks the deeper the transformer is. 

\begin{table}[htbp]
\caption{Table comparing T5-Small with T5-Base architectures}
\begin{tabular*}{\hsize}{@{\extracolsep{\fill}}lllll@{}}
\toprule
\textbf{Model}& \textbf{Embeddings Size} & \textbf{Attention Heads} & \textbf{Blocks} & \textbf{Parameters} \\
\colrule
T5-Base  & 768 &  12 & 12 & 220 M\\
T5-Small & 512  &  8 & 6 & 60 M\\
\botrule
\label{tab 55:t5 vs small t5 architecture}
\end{tabular*}
\end{table}

Table \ref{tab 5:t5 vs small t5} shows that the performance of T5-Small is similar to T5-Base with an accuracy drop of 1\% only on each dataset. In this case, reducing the size of the pre-trained model did not affect the results significantly, which could be attributed to the small size of both datasets indicating that having a large number of parameters is not necessary. The results also show that a small pre-trained language model is able to outperform transformers trained from scratch by a large margin. In addition, the large improvement in accuracy suggests that a relatively small pre-trained language model is able to acquire similar general knowledge during pre-training on language. Hence, having a smaller embedding size and a less number of attention heads and blocks does not prevent the model to generalize and retrieve the knowledge gained during pre-training.    

\begin{table}[htbp]
\caption{Table comparing the performance of T5-Small with T5-Base to study the effect of the size of the model on the performance}
\begin{tabular*}{\hsize}{@{\extracolsep{\fill}}lll@{}}
\toprule
\textbf{Model}& \textbf{Listops} & \textbf{CIFAR10-LRA} \\
\colrule
T5-Base  & \textbf{64.2\%} &  \textbf{58.5\%} \\

T5-Small & 63.4\%  &  57.6\% \\

Scratch Trans. & 37.3\%  &  44.2\% \\
\botrule
\label{tab 5:t5 vs small t5}
\end{tabular*}
\end{table}

\subsection{Can we get a similar performance with only 2\% of the parameters?}

After showing in the previous section that pre-trained models do not need a model with large number of parameters to improve the state-of-the-art by a large margin, in this section, inspired by FPT \cite{b10}, we freeze 98\% of the parameters and we fine-tune the remaining. In fact, we freeze all of the layers of T5-Small except for layer norm, which is a standard practice \cite{b25} \cite{b26}, and the output layer. Thus, the number of parameters is reduced from 60 to 2 millions.

We compare our work with FPT on three datasets and find that our model outperforms FPT with a large margin as shown in table \ref{tab 6:frozen t5}. The results show that even though the performance of frozen T5-Small drops significantly compared to fine-tuning the whole model, it still outperforms FPT \cite{b10} by a significant margin except for Remote Homology dataset where they both score the same. At first glance, it might seem that the accuracy on the Remote Homology dataset is low. However, the dataset has only around 12k training samples with 1195 classes to classify which averages to around 10 training samples per each class. Thus, the results are quite impressive keeping in mind that it is a biologic domain where the model has to predict the 3D protein fold given an amino acid sequence. The usual approach to tackle this dataset is to pre-train on the Pfam dataset \cite{b27}. The improvement on this dataset also shows how important pre-training on language is. Having only a small amount of data to be trained on does not prevent the model to improve the accuracy. It is another implication that general knowledge acquired during pre-training was retrieved even with small amount of data. 

In addition to not needing a lot of data, the results in table \ref{tab 6:frozen t5} show that the pre-trained models do not need all of the parameters to improve on a transformer that is fully trained from scratch. This would also suggest that the parameters learned during pre-training on language are also useful on tasks from different domains.  Thus, the results align with \cite{b10} where they claim that pre-trained transformers, which are minimally fine-tuned (only 2\% of the parameters in our case), are ``universal computation engines'' since they can achieve an equivalent performance, in their case and a better performance in our case, to a fully trained transformer from scratch. 

\begin{table}[htbp]
\caption{Table comparing the performance of T5-Small with T5-Base to study the effect of the size of the model on the performance}
\begin{tabular*}{\hsize}{@{\extracolsep{\fill}}llll@{}}
\toprule
\textbf{Model}& \textbf{Listops} & \textbf{CIFAR10-LRA}  & \textbf{Remote Homology}\\
\colrule
T5-Small & 63.4\% &  57.6\% & 13.9\% \\

Frozen T5-Small & 54.1\% &  45.7\% & 12.7\% \\

FPT & 38.4\%  &  38.6\% & 12.7\%\\

Scratch Trans. & 37.3\%  &  44.2\% & 9.0\% \\
\botrule
\label{tab 6:frozen t5}
\end{tabular*}
\end{table}

\subsection{Should we use pre-trained embedding input or reinitialize it?}

\begin{figure*}[htbp]
\begin{center}
\includegraphics[scale=0.6]{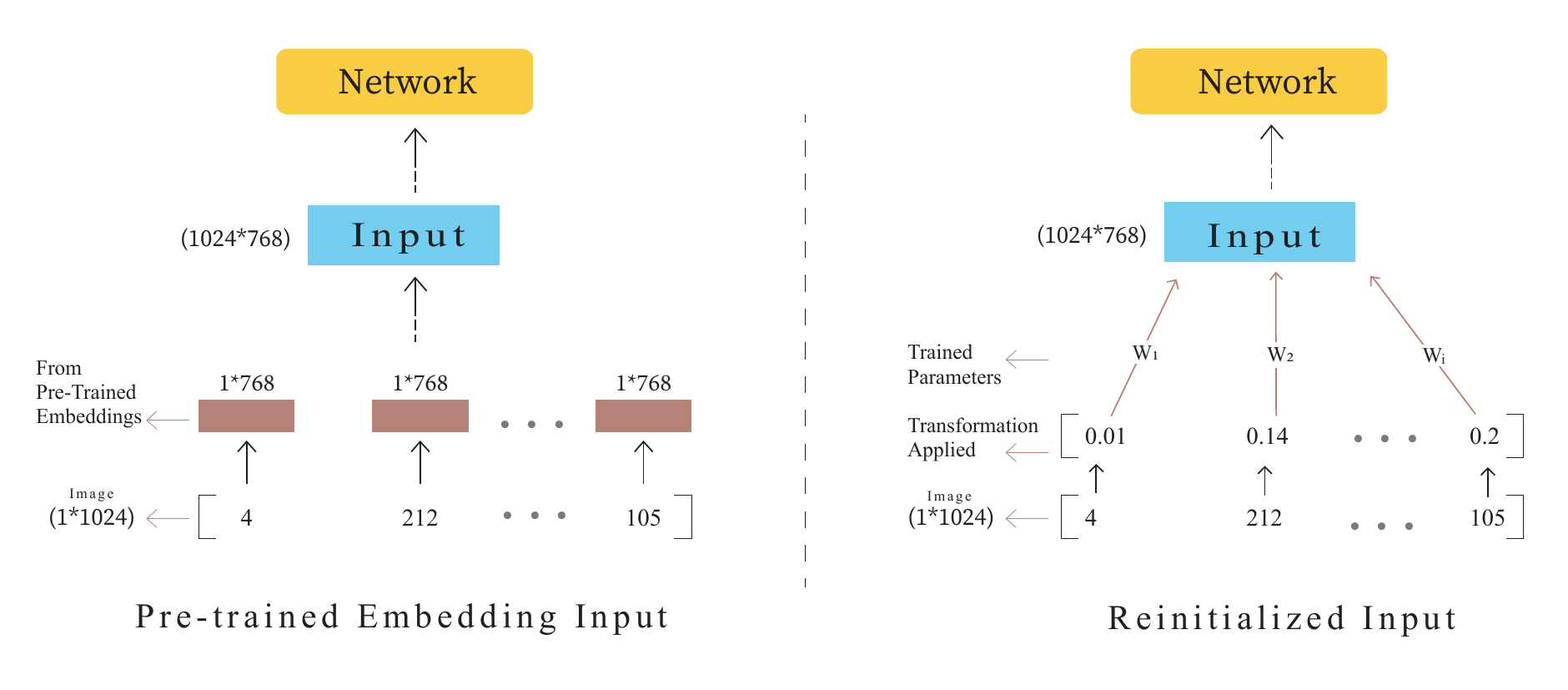} 
\caption{Figure shows the difference between having pre-trained embedding input and reinitialized input}
\label{fig:3}
\end{center}
\end{figure*}

In this section we compare our pre-trained GPT-2 with the pre-trained GPT-2 in \cite{b10} on the same datasets (Listops, CIFAR10-LRA, and Remote Homology). Table~\ref{tab 7} shows surprising differences.  
\begin{table}[htbp]
\caption{Table comparing our fully fine-tuned GPT-2 with fully fine-tuned GPT-2 from \cite{b10} on three datasets}
\begin{tabular*}{\hsize}{@{\extracolsep{\fill}}llll@{}}
\toprule
\textbf{Model}& \textbf{Listops} & \textbf{CIFAR10-LRA}  & \textbf{Remote Homology}\\
\colrule
GPT-2 \cite{b10}  & 35.8\% & 21.0\% & 10.5\% \\

Our-GPT-2  & 43.2\% & 62.0\% & 12.9\% \\
\botrule
\label{tab 7}
\end{tabular*}
\end{table}
The results in \cite{b10} show significantly lower performance after fine-tuning all parameters of GPT-2 for the specific tasks compared to our results. For instance, in the CIFAR10-LRA task, we achieved a 62.0\% accuracy compared to 21.0\% in \cite{b10}. Although our fine-tuning procedure is similar to that in \cite{b10}, there is a key distinction in how we handle the input layer. In \cite{b10}, a new input layer was created by randomly initializing its input weights, which were later fine-tuned. This is currently considered the standard approach when changing data modality. In contrast, we utilize pre-trained embeddings from the original model. In particular, we input the data into the network using the original word embeddings, which also generates embeddings for small integers. The difference between the two inputs is shown in figure \ref{fig:3}.

The results show that the pre-trained embeddings learned during pre-training on language have a crucial impact on the results. It suggests that the representations learned during pre-training are not exclusive for language, but they are useful on tasks from other domains as well. Hence, we claim that it is a good practice to transfer the pre-trained embeddings of the input even if the model is to be used on different domains.  

However, the drawbacks of using this method is that the only input allowed is integers because the pre-trained models represent integers with a vector of 768 points. On the other hand, the patches approach used by \cite{b10} represents each patch, which could be 16 pixels, by a vector of 768 points. The patch approach was created in order to make vision applications feasible for transformers, where input, for instance, could be $224 \times  224$. Thus, a good research direction is to work on making pre-trained embeddings feasible with the patches method. 

Thus, we compare our frozen T5 with the result reported for T5 by \cite{b10}. Our frozen T5 scores 54.1\% whereas frozen T5 from \cite{b10} scores 15.4\%. This also indicates a clear difference in performance. Note that we are only showing the results on the Listops dataset since it is the only score provided by \cite{b10} for T5.            

\section{Conclusion}

In this work, we investigated the ability of pre-trained language models to generalize to tasks from different domains. We fine-tuned four pre-trained models with different architectures and training styles on three non-language tasks. We showed that pre-trained language models are able to surpass the state-of-the-art trained from scratch transformers by a large margin when fully fine-tuned. T5, BERT, BART, and GPT-2 performed similarly on the non-language tasks showing that pre-training on language is the success factor. We also fine-tuned the models using 2\% of the parameters where the attention and feed-forward parameters are frozen, and the models outperformed transformers trained from scratch, but slightly worse than fully fine-tuned models. We also showed the importance of initializing the input with pre-trained embeddings as it was a key factor for the improvement.

This paper could serve as a base for future work regarding multi-task models and finding universal pre-training data. It could be extended by trying to test other pre-training techniques or data that could also create general knowledge representations. While the current research focuses on having specific pre-training data for each domain such as language, vision, etc., the ultimate goal would be to find a universal or combined data domains that could be general enough for cross-domain tasks.

\section*{Acknowledgements}

This work was funded by the Deutsche Forschungsgemeinschaft (DFG, German Research Foundation). The cluster used to train the models was also funded by the German Research Foundation (DFG) - 456666331.








\end{document}